\documentclass{article}
\usepackage{spconf,amsmath,amssymb,graphicx,hyperref}
\usepackage{float}
\usepackage{algorithm}
\usepackage{algpseudocode}

\def\tJ{\mathcal{J}}
\def\tX{\mathcal{X}}

\def\mA{\mathbf{A}}
\def\mB{\mathbf{B}}
\def\mD{\mathbf{D}}
\def\mG{\mathbf{G}}
\def\mH{\mathbf{H}}
\def\mM{\mathbf{M}}

\def\mR{\mathbf{R}}
\def\mS{\mathbf{S}}

\def\mV{\mathbf{V}}
\def\mW{\mathbf{W}}
\def\mX{\mathbf{X}}
\def\mY{\mathbf{Y}}

\def\va{\mathbf{a}}
\def\vb{\mathbf{b}}
\def\vc{\mathbf{c}}
\def\vf{\mathbf{f}}
\def\vg{\mathbf{g}}
\def\vh{\mathbf{h}}
\def\vr{\mathbf{r}}
\def\vu{\mathbf{u}}
\def\vv{\mathbf{v}}
\def\vw{\mathbf{w}}
\def\vx{\mathbf{x}}
\def\vy{\mathbf{y}}
\def\vz{\mathbf{z}}

\def\sR{\mathbb{R}}

\def\kr{\odot}
\def\tp{^{\intercal}}

\def\Neff{N_{\text{eff}}}

\DeclareMathOperator*{\argmin}{argmin}
\DeclareMathOperator{\diag}{diag}
\DeclareMathOperator{\tr}{tr}
\DeclareMathOperator{\GCV}{GCV}

\title{Making Tensor Decoupling More Robust and Less Parametric}

\name{Maxime Rochkoulets$^{1}$ \qquad Joppe De Jonghe$^{2}$ \qquad Mariya Ishteva$^{2}$}
\address{
  $^{1}$ KU Leuven, Belgium \qquad
  $^{2}$ NUMA, KU Leuven, Belgium
}

\begin{document}
\ninept
\maketitle

\begin{abstract}
Tensor decoupling algorithms recover compact representations of multivariate functions by combining linear factor matrices with univariate internal nonlinearities. We identify two limitations of recent methods. First, their performance is not invariant to mismatched output or gradient magnitudes, and second, their spline internals require manual tuning of their degrees of freedom. We propose two lightweight fixes. The first is a preprocessing step that rescales the input Jacobian tensor, removing both the bias towards larger magnitude outputs and the ill-conditioning of certain factor updates caused by such a mismatch. The second replaces B-spline internals by penalized splines coupled with generalized cross-validation, making the method less dependent on the user selecting the right number of control points. Finally, we release an efficient open-source JAX implementation of our algorithm.
\end{abstract}


\section{Introduction}
\label{sec:intro}

\subsection{Tensor Decoupling}

\emph{Tensor decoupling} \cite{dreesen2015decoupling} is a methodology for finding compact representations of multivariate functions. Given a target function $\vf$, decoupling algorithms look for an equivalent \emph{decoupled representation}
\begin{equation}\label{eq:decoupled}
\vf(\vx) \approx \mW \vg(\mV\tp \vx),
\end{equation}
in which a small number of univariate nonlinearities, collected in $\vg$ and called \emph{internal functions}, is sandwiched between two linear transformations $\mW$ and $\mV$. Such a representation is compact and possibly interpretable, since all the nonlinear behaviour of the target function is captured by a small set of one-dimensional curves \cite{dreesen2015wienerhammer, dreesen2018decoupling}.

More recent tensor decoupling algorithms \cite{hollander2017multivariate, zniyed2021tensor, de2025non} bring together several improvements over the original approach. Namely, they exploit the function values themselves (zeroth-order information) by coupling the tensor decomposition with a factorization of the output matrix, tie the last factor to the internal functions instead of fitting them afterwards, and represent the internals with more flexible models than polynomials, such as splines.

\subsection{Contributions}

Important limitations remain, and we identify two desirable properties that are not satisfied by current algorithms.

The first property is \emph{invariance to arbitrary rescaling of the target outputs or gradients}. In particular, we show that all current least-squares-based approaches are not suited for functions with highly mismatched output or gradient ranges (common in real-world problems) but that this can be easily fixed via a cheap preprocessing step.

The second is \emph{the ability for the algorithm to self-tune its expressivity through soft regularization}. The use of B-splines to model the univariate internals, while being a powerful and flexible approach, introduces a new hyperparameter which is non trivial to tune and easily leads to under- or overfitting. We show that adding a smoothing penalty to the spline internals, coupled with generalized cross-validation, leads to a penalized splines formulation that is \emph{less-parametric} with respect to the number of control points.

Both contributions are illustrated on synthetic examples, and an efficient open-source JAX \cite{jax2018github} implementation is released\footnote{\scriptsize\textsf{https://github.com/mrochk/decoupling}}, together with the source code to reproduce our experiments\footnote{\scriptsize\textsf{https://github.com/mrochk/ICASSP.2027}}.

\subsection{Notation}

Vectors, matrices and tensors are denoted by bold lower-case, bold upper-case and calligraphic letters, for example $\vx$, $\mX$ and $\tX$. For a third-order tensor $\tX$, we write $\tX_{(n)}$, $\tX_{::k}$ and $\tX_{i::}$ for its $n$-th mode unfolding, frontal slices and horizontal slices, respectively. The symbol $\kr$ denotes the Khatri-Rao product, and $\circ$ the outer-product.

\subsection{Canonical polyadic decomposition}

Given a tensor $\tX \in \sR^{m \times n \times p}$, a CP decomposition \cite{hitchcock1927expression, harshman1970foundations, carroll1970analysis} of $\tX$ aims to represent the tensor as a minimal sum of rank-one tensors
\begin{equation}\label{eq:cpd}
\tX \approx \sum_{i=1}^{r} \va_i \circ \vb_i \circ \vc_i = [\![\mA, \mB, \mathbf{C}]\!],
\end{equation}
with factor matrices $\mA \in \sR^{m \times r}$, $\mB \in \sR^{n \times r}$ and $\mathbf{C} \in \sR^{p \times r}$ collecting the component vectors as columns. The CPD \emph{rank} $r$ is the smallest value for which \eqref{eq:cpd} is an equality \cite{kolda2009tensor, TensorTextbook}. This decomposition is unique under mild conditions, and Kruskal's condition \cite{kruskal1977three} on the k-ranks of the factor matrices $\textnormal{k}_{\mA} + \textnormal{k}_{\mB} + \textnormal{k}_{\mathbf{C}} \geq 2r+2$ is the most well-known sufficient condition, although others exist \cite{sidiropoulos2017tensor}.

\subsection{The decoupling problem}

Let $\vf: \sR^m \to \sR^n$ be a multivariate vector-valued function. A \emph{single-layer decoupled representation} of $\vf$ is one for which \eqref{eq:decoupled} holds exactly, with $\mW \in \sR^{n \times r}$, $\mV \in \sR^{m \times r}$, and univariate internals collected in $\vg$ such that
\begin{equation}
\vg(\vu) = \begin{bmatrix} g_1(u_1) & g_2(u_2) & \dots & g_r(u_r)\end{bmatrix}\tp,
\end{equation}
for $\vu \in \sR^r$ with univariate functions $g_i: \sR \to \sR$. Solving the decoupling problem amounts to finding the matrices $\mW$, $\mV$, and representations of the internal functions $g_i$. Under this decoupled model assumption and via application of the chain rule, the Jacobian of $\vf$ evaluated at a point $\vx \in \sR^m$ is 
\begin{equation}\label{eq:jacobian}
\mathbf{J}(\vx) = \mW \diag(\vg'(\mV\tp \vx)) \mV\tp,
\end{equation}
with $\vg'(\vu) = \begin{bmatrix} g_1'(u_1) \; \dots \; g_r'(u_r)\end{bmatrix}\tp$ \cite{dreesen2015decoupling}. Evaluating \eqref{eq:jacobian} in $N$ points and stacking the resulting
matrices as frontal slices yields the Jacobian tensor
$\tJ \in \sR^{n \times m \times N}$ with CPD structure
\begin{equation}\label{eq:jcpd}
  \tJ = [\![\mW, \mV, \mH]\!] \quad \text{ with } \quad
  \mH_{j:} = \vg'(\mV\tp\vx_j)\tp.
\end{equation}
The solution strategy forming the basis of most decoupling methods can be summarized as
\begin{enumerate}
\itemsep0em
    \item[1)] evaluate the Jacobians and build the 3-way tensor $\tJ$,
    \item[2)] compute the rank-$r$ CPD of $\tJ$ to retrieve $\mW$, $\mV$ and $\mH$,
    \item[3)] recover the internal univariates $g_1 \, \dots \,g_r$.
\end{enumerate}
This approach, however, assumes uniqueness of the underlying CP decomposition, and is therefore not suited for dealing with so-called \emph{noisy CPD scenarios}.

\subsection{Coupled matrix-tensor B-spline decoupling (CMTF-BSD)}
\label{sec:cmtfbsd}

We briefly describe the CMTF-BSD approach \cite{de2025non}, used as a baseline in this work. Besides $\tJ$, this algorithm uses the corresponding inputs $\mX \in \sR^{N \times m}$ and outputs $\mY \in \sR^{N \times n}$. Zeroth-order information is exploited by coupling the CPD of $\tJ$ with a factorization of the output matrix $\mY \approx \mR\mW\tp$ \cite{acar2011all, zniyed2021tensor}, where the factor $\mR \in \sR^{N \times r}$ collects the (zeroth-order) evaluations of the internals. Then, $\mH$ and $\mR$ are constrained to be consistent with a common internal representation \cite{hollander2017multivariate}, where each $i$-th internal $g_i$ is modeled as a spline of degree $d$ evaluated at the projected samples $\vz_i = \mX \vv_i \in \sR^N$
\begin{equation}\label{eq:internalsplines}
\mH_{:i} = \mB_i' \vc_i, \qquad \mR_{:i} = \mB_i \vc_i,
\end{equation}
where $\mB_i', \mB_i \in \sR^{N \times (df+1)}$ are the B-spline design matrices (with an added column of zeros or ones \cite{de2025non}) evaluated at $\vz_i$, and $\vc_i \in \sR^{df+1}$ are the B-spline coefficients. The resulting problem
\begin{equation}\label{eq:cmtf}
\min_{\left(\mW, \mV, \mH, \mR\right)}
\left\Vert \tJ - [\![\mW, \mV, \mH]\!] \right\Vert^2
+ \gamma \left\Vert \mY - \mR\mW\tp \right\Vert^2 ,
\end{equation}
subject to \eqref{eq:internalsplines} with $\gamma \in \sR_{\geq 0}$ weighting the zeroth-order term, is solved by alternating least-squares. These constraints allow the algorithm to find sound representations even in noisy CPD scenarios.

\section{Output Scale Mismatch}
\label{sec:scaling}

\subsection{Unbalanced Jacobian magnitudes}

The first pathology is general to all existing least-squares-based decoupling algorithms. Namely, existing solutions do not account for the possibility of the Jacobian values of different component functions falling into \emph{very} unbalanced magnitude ranges. In terms of the tensor $\tJ$, we are considering targets whose horizontal slices have mismatched norms $\Vert\tJ_{i::}\Vert \gg \Vert\tJ_{j::}\Vert$ for some pair $(i,j)$. Note that this covers both a mismatch in output ranges and in gradients only.

The CMTF update of factor $\mW$ \cite{zniyed2021tensor} (the first update performed in each ALS iteration) derived from \eqref{eq:cmtf}, is given by
\begin{equation}\label{eq:updW}
\begin{split}
\mW \leftarrow \argmin_{\mW} & \left\Vert \tJ_{(1)} - \mW(\mH \kr \mV)\tp \right\Vert^2 \\ + \; \gamma & \left\Vert \mY\tp - \mW\mR\tp \right\Vert^2.
\end{split}
\end{equation}

Both targets $\tJ_{(1)}$ and $\mY\tp$ have rows indexed by output, and the design matrix is shared across rows. The problem therefore separates into $n$ independent row problems, such that the $i$-th row of the solution is scaled proportionally to the scaling of the $i$-th output, leading to the imbalance being absorbed in the row norms of $\mW$. For convenience, we write $\mW = \mS\bar{\mW}$, with $\mS = \diag(\Vert\mW_{1:}\Vert \; \dots \; \Vert\mW_{n:}\Vert)$ such that $s_{ii} \gg s_{jj}$ and $\bar{\mW}$ the row-normalized $\mW$.

\subsection{The output bias problem}

The most direct consequence of dealing with functions having mismatched magnitudes is that the algorithm becomes biased towards larger outputs. Looking at the other factor updates \cite{TensorTextbook, zniyed2021tensor}
\begin{align}
  \mV &\leftarrow \argmin_{\mV} \left\Vert \tJ_{(2)} - \mV(\mH \kr \mS\bar{\mW})\tp \right\Vert^2, \label{eq:updV}\\
  \mH &\leftarrow \argmin_{\mH} \left\Vert \tJ_{(3)} - \mH(\mV \kr \mS\bar{\mW})\tp \right\Vert^2, \label{eq:updH}\\
  \mR &\leftarrow \argmin_{\mR} \left\Vert \mY - \mR(\mS\bar{\mW})\tp \right\Vert^2, \label{eq:updR}
\end{align}
each of these splits into one independent subproblem per row of the unknown factor, and each such row is fit against a target spanning all $n$ outputs at once. The rows of $\tJ_{(2)}$ and $\tJ_{(3)}$ are indexed by input variable and by sample respectively, and each is a concatenation of $n$ segments (a length-$m$ Jacobian block for $\mV$ and $\mH$, a single scalar for $\mR$), whose $i$-th segment carries the scale $s_{ii}$ of the $i$-th output.

A single shared unknown is therefore fit against differently scaled quantities: the residual of the largest output dominates, and small magnitude outputs are in effect ignored.

\subsection{The conditioning problem}

The second issue caused by targets with vastly different output magnitudes is an increase of the condition number of the design matrix in the updates of $\mR$ (always), $\mH$ (when $m < r$), and $\mV$ (when $N < r$).

First, in the update of $\mR$ \eqref{eq:updR}, the design matrix is $\mW$ itself, whose rows are scaled by the corresponding factors in $\mS$. This always affects the condition number as soon as $n, r \geq 2$.

Now, consider the update of $\mH$ \eqref{eq:updH} with design matrix $\mA = (\mV \kr \mS\bar{\mW})$. Up to a row permutation, $\mA$ may be written as
\begin{equation}\label{eq:blocks}
  \mA =
  \begin{bmatrix}
    s_{11} \cdot \mV \diag(\bar{\vw}_1) \\ \vdots \\ s_{nn} \cdot \mV \diag(\bar{\vw}_n)
  \end{bmatrix}
  =
  \begin{bmatrix}
    s_{11} \cdot \mM_1 \\ \vdots \\ s_{nn} \cdot \mM_n
  \end{bmatrix},
\end{equation}
with $\mM_i \in \sR^{m \times r}$. The condition number of $\mA$ is
\begin{equation}\label{eq:kappa}
  \kappa(\mA) = \sqrt{\lambda^{\mA\tp\mA}_{\max}\, \big/\, \lambda^{\mA\tp\mA}_{\min}},
\end{equation}
where $\lambda^{\mA\tp\mA}_{\max}$ and $\lambda^{\mA\tp\mA}_{\min}$ denote the largest and smallest eigenvalue of $\mA\tp\mA = \sum_i s_{ii}^2 \cdot \mM_i\tp\mM_i$. We study how $\kappa(\mA)$ evolves with respect to $s_{\max} = \max(\mS)$ while the other scaling factors are fixed, we let $j$ be the index such that $s_{jj} = s_{\max}$.

For the largest eigenvalue, we directly have
\begin{equation}\label{eq:lmax}
  s_{\max}^2 \cdot \lambda^{\mM_j\tp\mM_j}_{\max}
  \;\leq\; \lambda^{\mA\tp\mA}_{\max}
  \;\leq\; s_{\max}^2 \cdot \sum_{i=1}^{n} \lambda^{\mM_i\tp\mM_i}_{\max},
\end{equation}
which yields $\lambda^{\mA\tp\mA}_{\max} = \Theta(s_{\max}^2)$. For $\lambda^{\mA\tp\mA}_{\min}$, we treat the $m \geq r$ and $m < r$ cases separately.

When $m \geq r$, $\mM_j$ generically has full column rank $r$, so that $\left\Vert \mM_j \vx \right\Vert^2 \geq \lambda^{\mM_j\tp\mM_j}_{\min} > 0$ for every unit vector $\vx$. Therefore we have
\begin{equation}\label{eq:lmin1}
  \lambda^{\mA\tp\mA}_{\min} \;\geq\; s_{\max}^2 \cdot \lambda^{\mM_j\tp\mM_j}_{\min},
\end{equation}
which combined with \eqref{eq:lmax} yields $\lambda^{\mA\tp\mA}_{\min} = \Theta(s_{\max}^2)$ and $\kappa(\mA) = \Theta(1)$. On the other hand, when $m < r$, which is common in practice, $\mM_j$ is column rank deficient. Picking a unit vector $\vx$ such that $\mM_j\vx = \mathbf{0}$, and denoting by $s_{(2)}$ the second largest scaling factor, we observe that
\begin{equation}\label{eq:lmin2}
\lambda^{\mA\tp\mA}_{\min} \;\leq\; \sum_{i \neq j} s_{ii}^2\,\left\Vert\mM_i\vx\right\Vert^2
\;\leq\; s_{(2)}^2 \cdot \sum_{i \neq j} \lambda^{\mM_i\tp\mM_i}_{\max},
\end{equation}
in other words, $\lambda^{\mA\tp\mA}_{\min}$ cannot exceed a fixed ceiling set by $s_{(2)}$. Since $\lambda^{\mA\tp\mA}_{\max} = \Theta(s_{\max}^2)$, the
condition number grows at least as fast as $s_{\max}\,/\,s_{(2)}$, and
$\kappa(\mA) \to \infty$ as $s_{\max} \to \infty$. 

Similar reasoning applies to the update of $\mV$ \eqref{eq:updV}, however, $N$ is typically chosen by the user to be larger than the rank \cite{dreesen2015decoupling}. The conditioning problem mainly affects the updates of $\mH$ and $\mR$.

\subsection{Solution: tensor rescaling}

A straightforward and cheap solution to both problems is to enforce
\begin{equation}\label{eq:equalnorms}
\left\Vert \tJ_{1::} \right\Vert \,=\, \left\Vert \tJ_{2::} \right\Vert \,=\, \cdots \,=\, \left\Vert \tJ_{n::} \right\Vert,
\end{equation}
that is, equality on the Frobenius norms of the horizontal slices of $\tJ$ (since $\tJ_{i::}$ contains all partial derivatives with respect to the $i$-th output). This suppresses the imbalance at its source, preventing it from being inherited by $\mS$. Requiring \eqref{eq:equalnorms} therefore solves both the output bias problem and the factors conditioning problem. It is easily achieved by rescaling $\tJ$ and $\mY$ as follows
\begin{equation}\label{eq:rescale}
  \tilde{\tJ}_{i::} \leftarrow \tJ_{i::} \cdot \frac{\sqrt{mN}}{\left\Vert \tJ_{i::} \right\Vert}
  \quad \text{ and } \quad
  \tilde{\mY}_{:i} \leftarrow \mY_{:i} \cdot \frac{\sqrt{mN}}{\left\Vert \tJ_{i::} \right\Vert},
\end{equation}
where the factor $\sqrt{mN}$ ensures that the norm does not shrink as the number of samples increases. Both the Jacobian tensor and the output matrix must be scaled by the same per-output factor, so that the coupling between the two terms is preserved. Also note that the representation returned by the algorithm must be unscaled before it can be compared to the target.

The Jacobian measurements used to compute the scaling factors in \eqref{eq:rescale} may themselves be noisy, leading to poor estimates. One alternative in this scenario is to estimate them using a larger set of $K > N$ samples, since the cost of this preprocessing is cheap.

Finally, this rescaling can be seen as a simple diagonal instance of the more general weighted formulation introduced in \cite{hollander2017multivariate}, which addresses the noisy setting, and in which the CPD residual is weighted by the inverse covariance of the Jacobian entries. General weights make it possible to encode per-output measurement noise levels or relative output importance, if such information is available.

\section{Spline Internals Smoothing}
\label{sec:smoothing}

\subsection{Degrees of freedom and overfitting}

CMTF-BSD uses B-splines to model the internal univariates. This introduces two hyperparameters: the internal spline degree $d$, which is commonly set to $d=3$ (cubic splines), and degrees of freedom $df$, defining the number of control points (number of basis functions).

It is tempting to increase $df$, as it maximizes the expressivity of the decoupled representation. However, B-splines with high $df$ values (close to $N$) tend to interpolate noisy points (especially in early iterations, after random initialization of the factors) and generalize poorly to unseen samples \cite{eilers1996flexible}. The question is therefore \emph{how to increase the number of control points while controlling for overfitting}.

\subsection{Solution: penalized splines}

Instead of B-splines, we use \emph{penalized splines} \cite{eilers1996flexible, eilers2015twenty} (P-splines) to represent our internals. This formulation adds a smoothing penalty term to the estimation of the spline coefficients
\begin{equation}\label{eq:pspline}
  \vc \leftarrow \argmin_{\vc}
  \left\Vert \vh - \mB'\vc \right\Vert^2
  + \gamma \left\Vert \vr - \mB\vc \right\Vert^2
  + \lambda \left\Vert \mD\vc \right\Vert^2,
\end{equation}
where $\vc \in \sR^{df+1}$ is the coefficient vector for one of the $r$ internals, $\vh \in \sR^{N}$ and $\vr \in \sR^{N}$ are the corresponding first and zeroth-order information (columns of $\mH$ and $\mR$) and $\gamma \in \sR_{\geq 0}$ is the CMTF weighting term. We introduce $\mD \in \sR^{(df-1) \times (df+1)}$, a second-order finite difference matrix\footnote{We found that second-order differences worked well in practice, but variations are possible \cite{eilers1996flexible, eilers2015twenty} and should be explored in future work.}, and $\lambda \in \sR_{\geq 0}$ is the weight of the smoothing term. Note that $\lambda = 0$ recovers the CMTF-BSD formulation.

\subsection{Generalized cross-validation}

At first glance, our P-spline formulation \eqref{eq:pspline} simply trades one hyperparameter ($df$) for another ($\lambda$). However, $\lambda$ can be tuned automatically and efficiently using generalized cross-validation (GCV) \cite{craven1978smoothing, golub1979generalized}. At each iteration and for each internal, we search for
\begin{equation}\label{eq:gcv}
\lambda^{*} \,=\, \argmin_{\lambda} \GCV(\lambda) \,=\, \argmin_{\lambda} \frac{\frac{1}{\Neff}\lVert \vy - \hat{\mB}\vc \rVert^2} {\big(1 - \frac{\tr(\mG(\lambda))}{\Neff}\big)^2},
\end{equation}
with $\vy = [\vh; \sqrt{\gamma}\,\vr]\tp$, $\hat{\mB} = [\mB'; \sqrt{\gamma}\,\mB]\tp$ and the \emph{hat-matrix} $\mG(\lambda) = \hat{\mB}(\hat{\mB}\tp\hat{\mB} + \lambda\mD\tp\mD)^{-1}\hat{\mB}\tp$. We empirically set $\Neff = N (1 + \gamma)$ as the \emph{effective number of observations}, accounting for the fact that zeroth-order rows of $\hat{\mB}$ are downweighted by $\gamma$.

This approach makes our method effectively non-parametric, or at least, \emph{less-parametric} with respect to the number of control points, since $df$ can now be set by default to a generous value $df \leq N$. The algorithm self-tunes its expressivity through the smoothing penalty, removing the need for extensive hyperparameter tuning of $df$.

\subsection{Implementation and cost}

This P-spline formulation introduces computational overhead. Indeed, a smoothing term $\lambda^*$ must be selected $r$ times for each of the $T$ iterations, and the cost of this search grows with both $df$ and $N$. A naive implementation (forming the hat-matrix $\mG(\lambda)$ and refitting \eqref{eq:pspline} for every candidate) would be prohibitively expensive.

Instead, we search over a grid of candidates in a Demmler-Reinsch-type basis \cite{demmler1975oscillation, ruppert2002selecting}, obtained from a pair of QR and singular value decompositions, in which the fitting and penalty terms are simultaneously diagonalized. In this basis, both the residual and $\tr\left(\mG(\lambda)\right)$ are available in closed-form for every candidate. In other words, the whole grid is scored for essentially the price of a single fit, and $\mG$ is never formed explicitly. Combined with JAX's JIT compiler \cite{jax2018github}, this makes the GCV search very fast in practice.

Concerning the knot placement, although evenly-spaced knots are more commonly used with P-splines \cite{eilers1996flexible, ruppert2002selecting}, we found quantile-based knot placement to be more stable and perform better in our experiments.

\section{Experiments}
\label{sec:experiments}

\subsection{Setup}

We now show both improvements separately using the following synthetic target function $\vf: \sR^2 \to \sR^2$
\begin{equation}\label{eq:target}
\vf(x_1, x_2; \sigma) =
\begin{bmatrix} \sigma & 0 \\ 0 & 1 \end{bmatrix}
\cdot
\begin{bmatrix}
\cos(4\pi x_1) + x_2^2 \\
-\sin(3\pi x_2) + x_1^3
\end{bmatrix},
\end{equation}
where varying the value of $\sigma \ge 1$ allows us to illustrate scaling problems. Most hyperparameters are common to all experiments, such as the rank which is set to $r=4$, the number of iterations $T=100$, the spline degree $d=3$, the CMTF weighting term $\gamma = 0.1$, and the number of random seeds which is set to $10$. We use $N=100$ random points sampled uniformly in $[0, 1]^2$.

\emph{Per-output error} is computed as follows
\begin{equation}
e_i = \frac{\sqrt{\frac{1}{N}\sum_{j=1}^N{(f_i(\vx_j) - \hat{f}_i(\vx_j))^2}}}{\sqrt{\frac{1}{N}\sum_{j=1}^N{\left(f_i(\vx_j) - \mathbb{E}_{\vx \sim \mX}[f_i(\vx)] \right)^2}}} \times 100,
\end{equation}
where $e_i$ denotes the error of the $i$-th output and \emph{mean per-output error} denotes the average of this metric taken over all random seeds. This metric is always computed on a test set $(\mX_{\text{test}}, \mY_{\text{test}})$, consisting of $N$ points sampled independently and unseen, to better evaluate the generalization capabilities of the returned representation.

\subsection{Output scale mismatch}

\begin{figure}[h]
  \centering
  \includegraphics[width=0.8\linewidth]{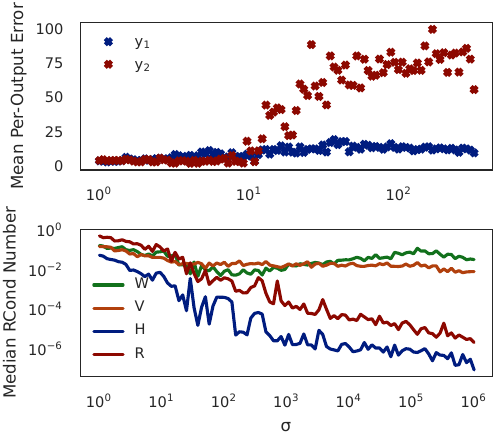}
  \vspace*{-12pt}
  \caption{Per-output errors and median reciprocal condition numbers when running CMTF-BSD \emph{without} rescaling, as a function of $\sigma$. Note that the two plots use different upper bounds for $\sigma$.}
  \label{fig:before}
\end{figure}

We first show the benefits of rescaling the input Jacobian tensor (Section \ref{sec:scaling}) by showing that when this preprocessing step is not applied (Figure \ref{fig:before}), increasing the value of $\sigma$ in \eqref{eq:target} leads to an increase of $e_2$ (bias problem) and to a decrease of the reciprocal condition number $1/\kappa$ for $\mR$ and $\mH$ (since $m\!=\!2 < r\!=\!4$) (conditioning problem). For the condition number, we report the median value over the $T$ iterations. In this experiment we used $df=14$ control points, and ranges of $100$ different values for $\sigma$.

\begin{figure}[h]
  \centering
  \includegraphics[width=0.8\linewidth]{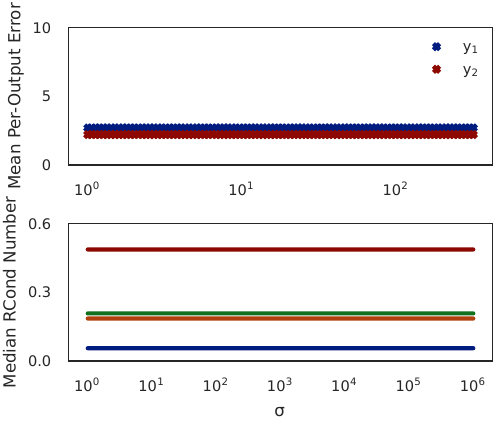}
  \vspace*{-12pt}
  \caption{Per-output error and median reciprocal condition numbers when running the CMTF-BSD algorithm \emph{with} tensor rescaling.}
  \label{fig:after}
\end{figure}

\noindent Applying the tensor preprocessing step fixes both the bias and the conditioning problem. This is illustrated in Figure \ref{fig:after}, where we perform the same experiment with the same hyperparameters and data, with $\tJ$ rescaled according to \eqref{eq:rescale} before running CMTF-BSD, effectively reaching the desired invariance to arbitrary rescaling.

\subsection{Internals smoothing}\label{sec:exp-smoothing}

\begin{figure}[h]
  \centering
  \includegraphics[width=0.9\linewidth]{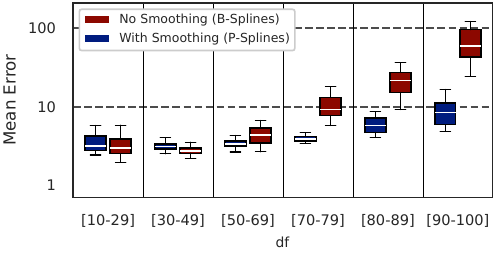}
  \vspace*{-12pt}
  \caption{Mean error box-plot (in log-scale) for $df = 10, 11\, \dots\, N$.}
  \label{fig:smoothing}
\end{figure}

To show the benefits of smoothing, that is, of replacing B-splines with P-splines coupled with GCV, we set $\sigma=1$ and vary the value of $df$ from $10$ to $N$. \emph{Mean error} in Figure \ref{fig:smoothing} is defined as $(e_1 + e_2)/2$, again averaged over all random seeds. For the GCV search, we use a fixed grid of 512 $\lambda$ values evenly spaced between $0$ and $10^{3}$. Results are grouped in buckets, allowing us to visualize the median and variance for similar $df$ values. When smoothing is not applied, the algorithm tends to overfit as $df \to N$, yielding poor representations, while splines with a smoothing prior degrade far more slowly. 

\vspace*{-15pt}

\section{Conclusion}

We identified two desirable properties for tensor decoupling algorithms: invariance to output or gradient rescaling, and the ability to tune the internals' expressivity in a non-parametric fashion. We addressed both with lightweight fixes. The first rescales the Jacobian tensor prior to decomposition, removing the bias and ill-conditioning that arise when outputs or gradients have mismatched magnitudes. The second replaces the B-spline internals of CMTF-BSD with P-splines tuned via GCV, making the algorithm less dependent on the user choosing the right number of control points. Both were shown to resolve their targeted pathology on a synthetic example, and an open-source JAX implementation of our algorithm is released.

Future work includes applying our algorithm to more complex real-world (e.g., nonlinear system identification \cite{decuyper2021retrieving}) problems, exploring alternatives for the smoothing penalty matrix, and investigating why quantile-based knot placement seemingly outperforms the equidistant placement more commonly used with P-splines.

\section{Acknowledgment}

This work was supported by the Fonds Wetenschappelijk Onderzoek (FWO) fundamental research fellowship 11A2H25N.

\bibliographystyle{IEEEbib}
\bibliography{refs}

\end{document}